\documentclass[conference]{IEEEtran}
\IEEEoverridecommandlockouts

\usepackage{dsfont}
\usepackage{amsfonts}
\usepackage{amsmath}
\usepackage{amssymb}
\usepackage{amsthm}

\usepackage{booktabs}
\usepackage{multicol}
\usepackage{multirow}
\usepackage{tabularx}
\usepackage{diagbox}
\usepackage{graphicx}
\usepackage{subfigure}
\usepackage{array}
\usepackage{fancyhdr}
\usepackage{hyperref}
 \hypersetup{
     colorlinks=true,
     linkcolor=blue,
     filecolor=blue,
     citecolor = black,      
     urlcolor=cyan,
     }
\usepackage{cleveref}
\usepackage{xcolor}
\usepackage{tikz}
\usepackage{verbatim}
\usepackage{listings}
\definecolor{mygreen}{RGB}{28,172,0} 
\definecolor{mylilas}{RGB}{170,55,241}

\usepackage{matlab-prettifier}
\usepackage{color}
\usepackage[shortlabels]{enumitem}

\usepackage{algorithm}
\usepackage[noend]{algpseudocode}
\makeatletter
\def\BState{\State\hskip-\ALG@thistlm}
\makeatother
\newcommand{\E}{\mathbb{E}}

\newcommand{\tp}{\mathsf{T}}

\newcommand{\N}{\mathbb{N}}
\newcommand{\R}{\mathbb{R}}

\newcommand{\IF}{\text{if}\quad}
\newcommand{\other}{\text{otherwise}\quad}

\newtheorem{lemma}{Lemma}

\newtheorem{proposition}{Proposition}

\newtheorem{definition}{Definition}

\def\BibTeX{{\rm B\kern-.05em{\sc i\kern-.025em b}\kern-.08em
    T\kern-.1667em\lower.7ex\hbox{E}\kern-.125emX}}

\usepackage{bbm}

\begin{document}

\title{Blackwell Online Learning for Markov Decision Processes
}

\author{\IEEEauthorblockN{ Tao Li}
\IEEEauthorblockA{\textit{Department of ECE} \\
\textit{New York University}\\
Brooklyn, USA \\
tl2636@nyu.edu}
\and
\IEEEauthorblockN{ Guanze Peng }
\IEEEauthorblockA{\textit{Department of ECE} \\
\textit{New York University}\\
Brooklyn, USA \\
gp1363@nyu.edu}
\and
\IEEEauthorblockN{ Quanyan Zhu }
\IEEEauthorblockA{\textit{Department of ECE} \\
\textit{New York University}\\
Brooklyn, USA \\
quanyan.zhu@nyu.edu}
}

\maketitle

\begin{abstract}
This work provides a novel interpretation of Markov Decision Processes (MDP) from the online optimization viewpoint. In such an online optimization context, the policy of the MDP is viewed as the decision variable while the corresponding value function is treated as payoff feedback from the environment. Based on this interpretation, we construct a Blackwell game induced by MDP, which bridges the gap among regret minimization, Blackwell approachability theory, and learning theory for MDP.  Specifically, from the approachability theory, we propose 1) Blackwell value iteration for offline planning and 2) Blackwell $Q-$learning for online learning in MDP, both of which are shown to converge to the optimal solution. Our theoretical guarantees are corroborated by numerical experiments.  
\end{abstract}


\section{Introduction}
Sequential decision making under uncertainty lies in the heart of many real world problems, ranging from portfolio management to robotic control. Many models and methods have been proposed to study the dynamic decision-making process, among which Markov decision process (MDP) \cite{puterman_mdp} and online optimization \cite{shai_online}, together with their variants, are most popular and well studied ones. Based on Bellman principle \cite{Bellman:1954uq}, solving an MDP problem relies on the backward induction using dynamic programming, where the future is considered when making decisions. For online settings, where the transition kernel and/or the reward function are unknown, reinforcement learning (RL) algorithms \cite{Sutton:2018wc} come into play, which more or less are based on dynamic programming idea. Combined with linear or nonlinear function approximators \cite{Tsitsiklis97TD,silver16go,tao_multiRL}, dynamic programming-based RL methods such as $Q-$learning \cite{Watkins:1992jx}, actor-critic \cite{konda99two} have brought about many empirical successes.      

 On the other hand, online optimization operates in a forward fashion, where decisions are made based on history and no information regarding the future is revealed during the process. In online optimization, the solution concept rests on the optimality in hindsight, widely referred to as no-regret property, as we shall introduce in the background. Closely related to the no-regret idea, Blackwell approachability \cite{blackwell56} gives a geometric interpretation of how regret vanishes over time in online decision-making. As pointed out in \cite{perchet14blackwell,abernethy11approach}, approachcbility and no-regret are equivalent, and we can develop no-regret algorithms based on the geometric intuition of approachability. As we will show in this paper, such connection can be made explicitly by considering a Blackwell game with vector-valued payoffs measuring the regret.  
 
Instead of studying MDP from an online optimization perspective, most prior works focus on online version of MDP, where transition and/or reward are time-varying, referred to as online MDP \cite{shimkin09onlinemdp,mansour09onlinemdp} or non-stationary RL \cite{NEURIPS2019_859b00ae,silva06}. Under this setting, the no-regret idea plays an important role, and we see no difficulty in extending our approachbility framework to these problems as we adopt the online learning viewpoint. Related to our work,  \cite{kash20no_regretQ} also applies the no-regret idea to MDP problems, which provides theoretical guarantees for offline settings. As shown in the paper, the convergence of the proposed method relies on no-absolute-regret algorithm, such as follow-the-regularized-leader (FTRL) with the linear cost. We argue that such no-regret method is a special case of our Blackwell approachability based framework.       
 
 In this paper, we take a step toward understanding MDP from the perspective of online optimization. We construct an auxiliary Blackwell game for MDP, so that we can leverage online optimization methods based on regret minimization. Our main contributions includes: 1) we give a no-regret value iteration algorithm, based on Blackwell approachability,  which we term  Blackwell value iteration. We show that this method provides asymptotic convergence guarantee as classical value iteration in discounted MDP; 2) We extend this idea to RL domain with unknown transition and reward, which accounts for online learning problems. Similar to $Q$-learning, our proposed method, Blackwell $Q$-learning, does not require any prior information nor any access to state sampling distribution. Hence, instead of an asynchronous version of value iteration \cite{Bertsekas1996NeurodynamicP}, our Blackwell $Q$-learning is indeed a RL algorithm based approachability idea. To the best of our knowledge, this is the first work that interprets an MDP as a Blackwell game, which leads to provably convergent learning algorithms.     
 
 The rest of the paper is organized as follows. We first introduce some preliminaries, including Blackwell approachability and no-regret in \cref{sec:back}. We then move to our proposed methods based on Blackwell approachability in \cref{sec:black}, where we give both value iteration like and $Q-$learning like algorithms for both offline planning and online learning problems. Our theoretical analysis is supported by numerical examples presented in \cref{sec:num}. Finally, we conclude the paper in \cref{sec:conclusion}. Due to the limit of space, we suppress all proofs in the paper and they can be found in the supplementary\footnote{Supplementary materials and code can be found at the repository site: \url{https://github.com/TaoLi-NYU/Blackwell-Online-Learning}.}.

\section{Background}\label{sec:back}

\subsection{Markov Decision Process}
An infinite-horizon discounted MDP can be characterized by a tuple, $\left\langle \mathcal{S},\mathcal{A}, \mathbb{P}, r, \gamma \right\rangle$,  where $\mathcal{S}$ is the finite state set; $\mathcal{A}$ is the finite action set; $\mathbb{P}:\mathcal{S}\times\mathcal{A}\rightarrow\Delta(S)$ is the transition probability and $\Delta(S)\subset \R^{|S|}$ denotes the simplex over $S$; $r:\mathcal{S}\times\mathcal{A}\rightarrow\mathbb{R}$ is the reward function; and $\gamma\in(0,1)$ is the discounting factor.

For a given policy $\pi:\mathcal{S}\rightarrow\Delta(\mathcal{A})$, the total expected reward starting from an initial state $s\in \mathcal{S}$ is defined as $V^\pi(s)=\E_{\mathbb{P},\pi}[\sum_{k=1}^\infty \gamma^k r(s_k,a_k)]$. If we denote $\pi(s,a)$ the probability of choosing $a$ at state $s$, then with the Bellman principle \cite{Bellman:1954uq}, $V^\pi$ can also be written as 
\begin{align*}
    V^\pi(s)=\sum_{a\in \mathcal{S}}\pi(s,a)\bigg[r(s,a)+\gamma\sum_{s'\sim \mathbb{P}(s,a)}V^\pi(s')\bigg],
\end{align*}
where we denote $Q^\pi(s,a)=r(s,a)+\gamma\sum_{s'\sim \mathbb{P}(\cdot|s,a)}V^\pi(s')$, known as the $Q$ function or $Q$ table. The  goal is to find an optimal policy $\pi^*$ such that $V^{\pi*}(s)\geq V^\pi(s)$ for all $s\in \mathcal{S}$. Similarly, for $\pi\in \R^{|S||A|}$, $\pi(s)\in \Delta(A)$.   

Since we focus on finite cases throughout this paper, all functions introduced above are of finite dimensions. To better present our work, we use the following notations. For  $Q\in\R^{|S||A|}$, $Q(s):=[Q(s,a)]_{a\in \mathcal{A}}$ denotes the vector in $\R^{|A|}$. Similarly, for $\pi\in \R^{|S||A|}$, $\pi(s)$ denotes the a vector in $\Delta(\mathcal{A})$. Finally, we assume that for every $s\in \mathcal{S}$ there exists an action $a\in \mathcal{A}$ such that the Markov chain is aperiodic and irreducible, which is a common assumption in reinforcement learning \cite{Sutton:2018wc}.

\subsection{Blackwell Approachability }
Blackwell approachability theory \cite{blackwell56} was developed for studying repeated game play between two players with vector-valued payoffs. In such a game, which we refer to as \textit{Blackwell game}, at the $k$-th round, both Player 1 and Player 2 select their actions  $x_k\in \mathcal{X}$ and $y_k\in \mathcal{Y}$ and then player 1 incurs the vector-valued payoff given by $u(x_k,y_k)\in \R^m$, where $u:\mathcal{X}\times\mathcal{Y}\rightarrow\R^m$ is a bi-affine function. We assume that action sets $\mathcal{X}, \mathcal{Y}$ are compact and convex. The objective of Player 1 is to guarantee that the average payoff converges to a desired closed convex target $\mathcal{D}\in \R^m$. We let $d(x,\mathcal{D}):=\inf_{z\in \mathcal{D}}\|x-z\|$ denote the distance between a point $x\in\R^n$ and the set $\mathcal{D}$ under norm $\|\cdot\|$. If we consider the Blackwell game $\left\langle\mathcal{X},\mathcal{Y}, u, \mathcal{D}\right\rangle$, then we can define an approachable set for Player 1 as follows.
\begin{definition}[Approachable Set \cite{blackwell56}]
	A set $\mathcal{D}$ is said to be approachable for Player 1, if there exists an algorithm $\sigma_k(\cdot):\mathcal{X}^{k}\times\mathcal{Y}^{k}\rightarrow\mathcal{X}$ which chooses an action at each round based on the history of play: $x_k=\sigma_{k}(x_{0:k-1},y_{0:k-1})$, such that for any sequence of $\{y_k\}_{k=1}^K$, $\lim_{K\rightarrow\infty}d(\frac{1}{K}\sum_{k=1}^Ku(x_k,y_k),\mathcal{D})=0$.
\end{definition}
    
A key concept in Blackwell approachability theory is the approachable halfspace, defined as below.
\begin{definition}[Approachable Halfspace \cite{blackwell56}]
	A halfspace  $\mathcal{H}:\{z \in \R^m| a^\tp z\leq b\}$ for some $a\in \R^m, b\in \R$ is approachable for Player 1 if there exists $x^*\in \mathcal{X}$ such that for all $y\in \mathcal{Y}$, $u(x^*,y)\in \mathcal{H}$.
\end{definition}
Blackwell's approachability theorem states that $\mathcal{D}$ is approachable if and only if all halfspaces $\mathcal{H}$ that contain $\mathcal{D}$ are approachable. Based on this theorem, we can construct a Blackwell strategy that guarantees the approachability, as shown in \cite{blackwell56}. 

\begin{figure}
    \centering
    \includegraphics[width=0.45\textwidth]{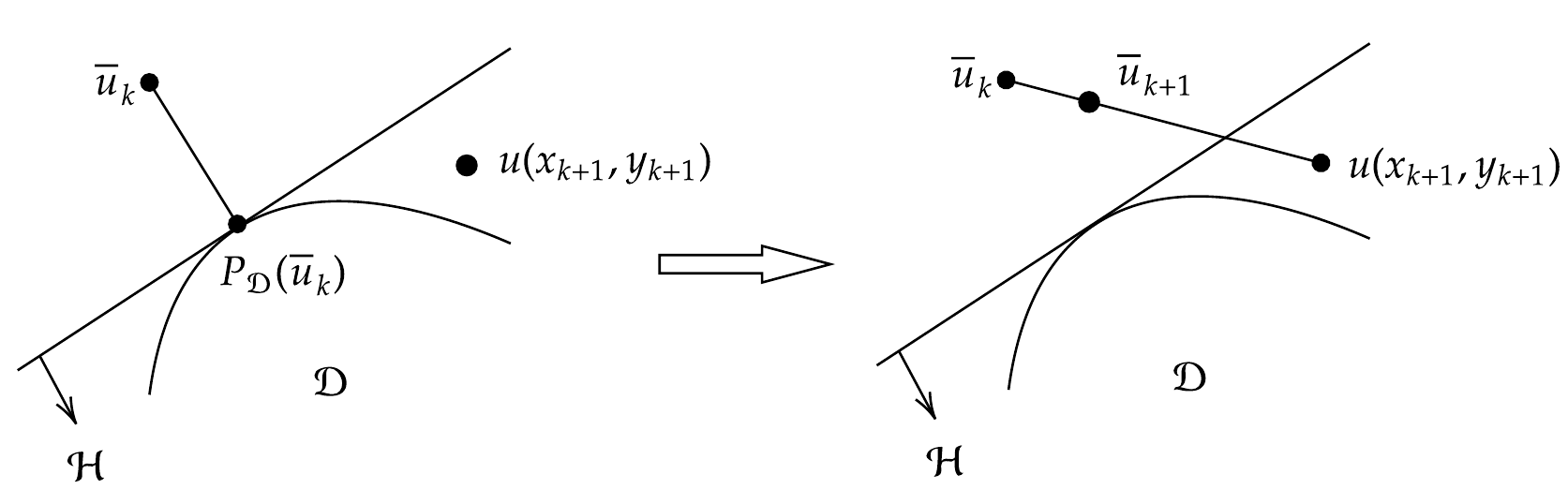}
    \caption{Blackwell strategy ensures that the next iterate $u(x_{k+1},y_{k+1})$ always falls within the halfspace $\mathcal{H}$, no matter what $y_{k+1}$ is. }
    \label{fig:approach}
\end{figure}

We denote the average payoff up to time $k$ by $\bar{u}_k:= \sum_{i=1}^k u(x_i, y_i)/k$  and the projection operator regarding the set $\mathcal{D}$ by $P_\mathcal{D}(x):=\{z\in \mathcal{D}:\|z-x\|=d(x,\mathcal{D})\}$. Since we  deal with convex sets, $P_\mathcal{D}(x)$ returns a singleton. If $\mathcal{D}$ is approachable, then the halfspace $\mathcal{H}$ defined by $\mathcal{H}:=\{z:\left\langle z, \bar{u}_k-P_\mathcal{D}(\bar{u}_k)\right\rangle\leq 0\}$ is approachable. Therefore, there exists $x^*\in\mathcal{X}$ such that for all $y$, $u(x^*,y)\in H$ and hence, if we let $x_{k+1}=x^*$, $u(x_{k+1},y_{k+1})$ falls into the same halfspace as the set $\mathcal{D}$ does. By doing so, we make $\bar{u}_{k+1}$ closer to the set, as shown in Fig.~\ref{fig:approach} and repeating the same procedure at each round, the average payoff converges to $\mathcal{D}$. 


To better present our idea of leveraging Blackwell approachability and no-regret idea to solve RL problems,  we first consider an example of online learning, which is a repeated game between the player and the nature. At each time $k$, the player chooses an action $x_k\in\Delta^m\subset\R^m$, a simplex in $\R^m$, while the nature chooses a payoff vector $y_k\in \R^m$, which evaluates the action according to the revealed payoff $\left\langle x_k,y_k\right\rangle$. Here, $m$ is a positive integer. Then, the regret for not having played action $e_i\in \Delta^m$ at time $k$ is given by $y_k(i)-\left\langle x_k,y_k\right\rangle$, measuring the difference of counterfactual outcomes of $e_i$ and the received payoff, where $y_k(i)$ is the $i$-th element of the vector $y_k$. Naturally, one would like to have a sequence of $\{x_k\}_{k}$ that achieves the best possible result: 
\begin{align}\label{eq:noreg}
    \lim_{K\rightarrow\infty}\frac{1}{K}\max_{x\in \Delta^m}\sum_{k=1}^K (\langle x,y_k\rangle-\langle x_k, y_k\rangle) =0,
\end{align}
showing that the sequence yields the same average performance as the best action in hindsight and a sequence is said to achieve no regret if it satisfies \eqref{eq:noreg}. One way to construct such no-regret sequences is to leverage Blackwell approachability as we show in the following.  

We consider the Blackwell game $\left\langle\Delta^m, \R^m, u, \R^m_{-} \right\rangle $, where $u:\Delta^m\times\R^m\rightarrow\R^m$ is 
\begin{equation*}
\begin{aligned}
    &\left({x}_{k}, {y}_{k}\right) \mapsto y_k-\left\langle {x}_{k},{y}_{k}\right\rangle \mathbf{1}_m\\
    &\qquad\qquad\qquad=\left(y_{k}(1)-\left\langle{x}_{k}, {y}_{k}\right\rangle, \ldots,y_{k}(m)-\left\langle{x}_{k}, {y}_{k}\right\rangle\right),
\end{aligned}
\end{equation*}
where $\mathbf{1}_m\in\mathbb{R}^m$ is an all-ones vector. We note that such $u$ measures the change in regret incurred at time $k$.  If we adopt Blackwell strategy, we aim to find $x\in\Delta^m$ such that for all $y$,
$$\left\langle u(x,y), \bar{u}_k-P_{\R^m_{-}}(\bar{u}_k)\right\rangle=\left\langle u(x,y), [\bar{u}_k]^+\right\rangle\leq 0,$$ where $([u]^{+})_i:=\max\{u_i,0\}.$ If we let 
\begin{align}\label{eq:rm}
\mathcal{RM}(x_{1:k},y_{1:k})=\left\{
	\begin{aligned}
		&[\bar{u}_k]^+/\|[\bar{u}_k]^+\|_1,  \IF[\bar{u}_k]^+\neq0 \\
		& \text{any point in } \Delta^m, \other  
	\end{aligned}\right.\tag{\text{RM}}
\end{align}
then we obtain that, for $x_{k+1}=\mathcal{RM}(x_{1:k},y_{1:k})$,
\begin{align*}
	\left\langle u(x_{k+1},y), [\bar{u}_k]^+\right\rangle &=\left\langle y-\left\langle x_{k+1},y\right\rangle\mathbf{1},[\bar{u}_k]^+ \right\rangle\\
	&=\left\langle y , [\bar{u}_k]^+\right\rangle-\left\langle[\bar{u}_k]^+ ,y\right\rangle=0,
\end{align*}
showing that \eqref{eq:rm} is indeed a Blackwell strategy. Intuitively, this strategy outputs the next action $x_{k+1}$ that is proportional to current cumulative regret $\bar{u}_k$: actions with larger regret shall be player more frequently, as they bring up better payoffs. Hence, it is also referred to as regret matching (RM) and has been studied in various contexts, including game theory \cite{hart00regret_match, hart03regret_cont_time} and online optimization \cite{perchet14blackwell}.

\section{Blackwell $Q$-learning}\label{sec:black}
In this section, we present how to incorporate Blackwell approachability framework into MDP problem through the Blackwell game we introduced above.
\subsection{Blackwell Value Iteration: Offline Planing }
We first address the planing problem of MDP, where the transition kernel and reward function is known. From dynamic programming perspective, to solve such a MDP problem, we either resort to value iteration or policy iteration, both of which relies on the stationarity of the environment. However, as we have mentioned before, online learning (optimization) methods operate in a forward fashion for non-stationary or time-varying systems.  In this subsection, we show that under stationary environments, online learning methods also guarantee the optimality of the solutions.

As for implementation, we first initialize a $Q$-value table, $Q_0$, and an initial policy $\pi_0(s)$ for each $s\in\mathcal{S}$. We run $|\mathcal{S}|$ copies of the algorithm; i.e., one for each state $s\in\mathcal{S}$, and we iteratively reveal the rewards $r(s,a)$ for all $a\in \mathcal{A}$, which further translates to the payoffs in the online learning problem induced by MDP. In this algorithm, for each state $s\in\mathcal{S}$, we view the policy $\pi_k(s)\in \Delta(\mathcal{A})$ as the decision variable and $Q_k(s)=(Q_k(s,a))_{a\in \mathcal{A}}$ as the payoff vectors, which is obtained by expected SARSA \cite{seijen09SARSA}
\begin{align}\label{eq:expectedSARSA}
Q_{k}(s,a)=r(s,a)+\gamma\E_{s'\sim \mathbb{P}(\cdot|s,a), a'\sim \pi_{k-1}}[Q_{k-1}(s',a')]	.
\end{align}
In this case, the payoff of the decision $\pi_k(s)$ is given by $\left\langle \pi_k(s),Q_k(s) \right\rangle$.
     
Similar to the Blackwell game in \cref{sec:back}, we can also construct an approachability game for MDP. For the decision $\pi_k(s)$ and the feedback $Q_k(s)$, we define  cost $\mathcal{R}:\Delta(\mathcal{A})\times \R^{|A|}\rightarrow\R^{|A|}$ as
\begin{align*}
	(\pi_k(s),Q_k(s))\mapsto Q_k(s)-\left\langle \pi_k(s),Q_k(s)\right\rangle\mathbf{1}_{|A|}.
\end{align*}
We note that for a given $\pi, Q$, the $i$-th entry of $\mathcal{R}(\pi(s),Q(s))$ measures the quality of the policy $\delta(a_i)$, i.e., simply choosing $a_i$, compared with current policy $\pi$. Intuitively, larger $\mathcal{R}_i$ implies that $\delta(a_i)$ could have given a better payoff, had it been implemented. 

It is not surprising that when using the Blackwell strategy such as \eqref{eq:rm}: $\pi_{k+1}=\mathcal{RM}(\pi_{0:k}, Q_{0:k})$, we drive the averaged regret $\bar{\mathcal{R}}_n=\frac{1}{n}\sum_{k=0}^{n-1}\mathcal{R}(\pi_k(s),Q_k(s))$ to the non-positive orthant $\R^{|A|}_{-}$ and in the limit; i.e.,  no action can produce a positive regret, showing that the limiting point achieves optimality. Since we consider the average regret under the Blackwell framework, our convergence result in \cref{prop:offline} is also about the average $Q$ tables.
\begin{proposition} \label{prop:offline}
        Let $\bar{Q}_n=1/n \sum_{k=0}^n Q_k$ and $Q^*$ be the $Q$ tables under the optimal policy, then 
        $\lim_{n\rightarrow\infty}\bar{Q}_n=Q^*.$
\end{proposition}
 Similar result has also been shown in \cite{kash20no_regretQ}, when applying FTRL. Compared with their approach, our Blackwell approachability-based method is in fact more generic. We argue that for the linear cost considered in that paper i.e., $\left\langle \pi(s), Q(s) \right\rangle$, it can be shown that FTRL is equivalent to RM proposed here. The details are included in the supplementary, which is mainly based on the connection between Blackwell approachability and online linear optimization studied in \cite{abernethy11approach}. On the other hand, though RM is probably the most natural Blackwell strategy, it is definitely not the only one. In the supplementary, we claim that various online linear optimizers, including online gradient descent and mirror descent algorithms, all can be leveraged to construct Blackwell strategies, offering much freedom in designing algorithms.
\subsection{Blackwell Q-learning: Online Learning}
Though intuitive and provably convergent, Blackwell value iteration only applies to the  offline setting where full information regarding the MDP is known. However, in standard RL problems, the agent is required to find the optimal policy without any access to the transition probability and reward functions. Hence, we need an online version of such approachability-based algorithms. 

As pointed out in \cite{kash20no_regretQ}, developing such online learning schemes is not straightforward, and there are two major challenges. The first one is about rewards revelation: in the Blackwell value iteration, we require that at each iteration, rewards  $r(s,a)$ for all actions $a\in \mathcal{A}$ at a certain state $s\in \mathcal{S}$ are revealed for updating the $Q$-table according to \eqref{eq:expectedSARSA}. However, in RL, since the reward function is unknown, we only have access to the feedback corresponding to the actual action executed at each iteration. Apparently, such bandit feedback cannot produce the regret vector $\bar{\mathcal{R}}_k$. One workaround proposed in \cite{kash20no_regretQ} is to use importance sampling technique in multi-arm bandit problems \cite{slivkins19bandit}. The update rule becomes $Q_{k+1}(s,a)=\left(r(s,a)+\gamma\E_{\pi_k}[Q_k(s',a')]\right)/\pi_{k}(s,a)$ if $a$ is an action sampled from $\pi_k(s)$ for state $s$, and $Q_{k+1}(s,a)=0$ otherwise. Unfortunately, as in bandit problems, the incorporation of importance sampling can only ensure expectation convergence, a weaker guarantee than almost sure convergence in $Q$-learning, which is less desirable in practice.     

The other challenge is that the asynchronous update in online learning is more involved than the synchronous one. In Blackwell value iteration, we update every state at every iteration, whereas in an online setting, this synchronous update is impossible, which introduces additional complexity to the convergence analysis. Different from the synchronous update, in the current iteration at time $k$, $Q_k(s)$ may be updated at different time instances for different states for the first $k$ steps. And it is highly likely that some states are visited more frequently than others. One straightforward remedy is to require all states to be visited with the same frequency, i.e., the state to be updated at each iteration is chosen uniformly from $\mathcal{S}$. With this additional condition, the asynchronous version of value iteration still guarantees the convergence as shown in \cite{kash20no_regretQ}, though it still falls within the realm of offline planning as the state transition, instead of fixed, is influenced by the chosen actions in online settings.       

In this subsection, we propose an online learning scheme based on the Blackwell approachability. We address these issues by the two-time scale asynchronous stochastic approximation \cite{Borkar:2009ts}.  By leveraging the Lyapunov stability theory of differential inclusion developed in \cite{benaim05SADI,benaim06SADI}, we show that adopting the Blackwell strategy in asynchronous update gives a provably convergent online learning scheme for tackling RL problems.

As we have discussed above, in the online setting, the state transition is influenced by executed actions, sampled from the policy, which further influences the update of the $Q$-table. This coupled dynamics of the policy update and the $Q$-table update makes it difficult to directly extend value iteration to online learning. One way to decouple the two dynamics is to adjust the timescales of the two. Simply put, we update the $Q$-table in the faster timescale while the policy in the slower one, where the faster timescale sees the slower as quasi-static while the slower timescale sees the faster as equilibrated. Specifically, we consider the following learning scheme based on the regret matching we have introduced. 

Let $s_{k+1}\in \mathcal{S}$ and $a_{k+1}\in\mathcal{A}$ be the state and action visited at time $k+1$, and the agent receives a noised reward $R_{k+1}$ from the environment. We assume that $R_{k+1}$ is unbiased in the sense that for the $\sigma$ fields $\mathcal{F}_{k+1}=\sigma\{a_{0:k+1},s_{0:k+1}, R_{0:k+1}\}$, $\mathbb{E}[R_{k+1}|\mathcal{F}_{k+1}]=r(s_{k+1},a_{k+1})$. Since states and actions are visited asynchronously, the agent need the asynchronous counters $\phi_k(s,a):=\sum_{i=1}^k\mathbbm{1}_{\{(s_i,a_i)=(s,a)\}}, \psi_k(s):=\sum_{i=1}^k\mathbbm{1}_{\{s_i=s\}}$, together with the step sizes $\{\alpha(k)\}_{k\in \N}, \{\beta(k)\}_{k\in \N}$, to determine the learning rates. Based on all the above, the agent estimates the $Q$ function by \eqref{onpolicy} and then updates its policy by \eqref{emp_freq}. Finally, the agent chooses an action based on regret matching idea in \eqref{eq:rm}, i.e, sampling an action from the probability proportional to $[ \mathcal{R}(\pi_{k}(s), Q_{k}(s))]^{+}$, which we denote by  $\mathcal{RM}(\pi_{k}(s),Q_{k}(s))$ with abuse of notations. We summarize the scheme in the following: for every $s\in\mathcal{S}$ and $a\in\mathcal{A}$, 
\begin{align}
    &Q_{k+1}(s,a)=Q_k(s,a)+\alpha(\phi_{k+1}(s,a))\nonumber\\
	&\qquad\cdot\mathbbm{1}_{\{(s,a)=(s_{k+1},a_{k+1})\}}[R_{k+1}+\gamma V_k(s_{k+2})-Q_k(s,a)],\label{onpolicy}\\
	&\pi_{k+1}(s)=\pi_{k}(s)+\beta(\psi_{k+1}(s))\mathbbm{1}_{\{s=s_{k+1}\}}[e_{a}-\pi_k(s)],\label{emp_freq}\\
	& a_{k+1}\sim \mathcal{RM}(\pi_{k}(s),Q_{k}(s)),\label{sampling}
	\end{align}
where $V_k(s)=\sum_{a\in \mathcal{A}}\pi_k(s,a)Q_k(s,a)$, and $e_a$ is the unit vector in $\R^{|A|}$.  We note that different from Blackwell value iteration, we here do not rely on all historical $\pi_k$ and $Q_k$ as our stochastic approximation schemes \eqref{onpolicy} and \eqref{emp_freq} already return averaged results. 

It is clear that \eqref{onpolicy} and \eqref{emp_freq} are coupled as $V_k$ involves both $\pi_k$ and $Q_k$. Technically speaking, in order to analyze the limiting behavior of the coupled dynamics, we must ``decouple'' them, and one possible approach as proposed in \cite{Borkar:2009ts,konda99two} is to adjust the timescales. Specifically, in our case, we require that $\beta(k)=o(\alpha(k))$, meaning that the $Q$ update \eqref{onpolicy} operates at a faster timescale than the policy one \eqref{emp_freq}. Intuitively speaking, when synchronous update \eqref{eq:expectedSARSA} becomes impossible in an online setting, in order to produce a feedback $Q_k$ that can approximately evaluate the current policy $\pi_k$, we must wait until $Q_k$ stabilizes before we update the policy. By running the policy update at a slow timescale, $Q$ updates see $\pi_k$ as quasi-static, hence \eqref{onpolicy} can be viewed as expected SARSA \cite{seijen09SARSA}, while policy update sees $Q_k$ as stabilized, serving as an approximation to $Q^{\pi_k}$.  In the subsequent, we show that the two timescale stochastic approximation indeed converges to the optimal $Q$ function and policy.       
\subsubsection{Convergence of the fast timescale}
In order to solve for the $Q$-learning problem defined in \eqref{onpolicy}, we resort to stochastic approximation introduced in \cite{benaim05SADI,benaim06SADI} and we first rewrite it in a more concise form. We define an operator $\mathcal{T}:\R^{|S||A|}\times\R^{|S||A|}\rightarrow\R^{|S||A|}$ whose $(s,a)$ entry is given by  $\mathcal{T}_{(s,a)}(\pi,Q):=r(s,a)+\gamma\sum_{a\in \mathcal{A}}\pi(s,a)Q(s,a)$ and then we define a vector $\Gamma_{k+1}\in \R^{|S|}$, whose $s$ entry is 
\begin{equation*}
    \begin{aligned}
        &\Gamma_{k+1}{(s)}\\
        &\quad=\mathbbm{1}_{\{s=s_{k+1}\}}[R_{k+1}+\gamma V_k(s_{k+2})-\mathcal{T}_{(s_{k+1},a_{k+1})}(\pi_k,Q_{k})].
    \end{aligned}
\end{equation*}
Note that \eqref{onpolicy} is equivalent to 
\begin{equation*}
    \begin{aligned}
    &Q_{k+1}(s,a)-Q_k(s,a)=\alpha(\phi_{k+1}(s,a))\mathbbm{1}_{\{(s,a)=(s_{k+1},a_{k+1})\}}\\
    &\qquad\qquad\qquad\qquad\cdot[\mathcal{T}_{(s,a)}(\pi_k,Q_k)-Q_k(s,a)+\Gamma_{k+1}{(s)}].
    \end{aligned}
\end{equation*}
We further define the asynchronous step sizes $\bar{\alpha}_k$ and the relative step sizes $\mu_k(s,a)$ as 
$$\bar{\alpha}_k:=\max_{(s,a)}\alpha(\phi_k(s,a)), \qquad \mu_k(s,a):={\alpha(\phi_k(s,a))}/{\bar{\alpha}}.$$
By letting $M_k$ be the $|S||A|\times |S||A|$ diagonal matrix whose $(s,a)$ entry is given by $\mu_k(s,a)$, we can rewrite the asynchronous update \eqref{onpolicy} as 
\begin{equation*}
    Q_{k+1}-Q_k=\bar{\alpha}_{k+1}M_{k+1}[\mathcal{T}(\pi_k,Q_k)-Q_k+\Gamma_{k+1}\otimes\mathbbm{1}_{|A|}],
\end{equation*}
where $\otimes$ denotes Kronecker product.

Denote the interpolated version of the stochastic approximation of the update above by $\{\bar{Q}_t(s)\}_{s\in\mathcal{S}}$ (See Definition 2.2 in \cite{Perkins12asy_SA}), where $t$ is the continuous time index.

\begin{proposition} \cite{Perkins12asy_SA}
There exists $0<\eta<1$ such that almost surely, the interpolated stochastic approximation $\{\bar{Q}_t(s)\}_{s\in\mathcal{S}}$ is an asymptotic pseudo-trajectory to the differential inclusion,
\begin{equation}\label{onpolicy_diff}	
    {d}{Q}_t(s)/{dt}\in \Omega^{\eta}[\mathcal{T}_{s}(\pi_t, Q_t)-Q_t], 
\end{equation}
where 
\begin{equation}\label{omega_eta}
    \Omega^{\eta}:=\left\{\operatorname{diag}(\{\omega(s)\}_{s\in\mathcal{S}}):\omega(s)\in[\eta,1],\forall\ s\in\mathcal{S}\right\}.
\end{equation}
\end{proposition}

\begin{proposition}
	When $\pi$ is fixed, $Q^\pi$ is the unique global attractor of \eqref{onpolicy_diff}.
\end{proposition}


\subsubsection{Convergence of the slow timescale} 
From the discussions above, the sequence of $Q$-tables $\{Q_k\}_k$ converge to the $Q^{\pi}$ when $\pi$ is fixed and it are Lipschitz continuous in $\pi$ \cite{Perkins12asy_SA}. Therefore, we can study the limiting behavior of \eqref{emp_freq} by analyzing its continuous counterpart in \eqref{emp_freq_diff}, where we can replace the $Q$-table in \eqref{emp_freq} with the attractor $Q^{\pi_t}$ given the current policy $\pi_t$, as the slow timescale views the $Q$ updates as stabilized. 
\begin{equation}\label{emp_freq_diff}
    {d}{\pi}_t(s)/{dt}\in \Omega^{\eta'}[\mathcal{RM}(\pi_t,Q^{\pi_t})-\pi_t(s)],
\end{equation}
where $\Omega^{\eta^\prime}$ is defined similarly as \eqref{omega_eta}. It remains to show that the differential inclusion \eqref{emp_freq_diff} has a global attractor, which we prove by standard Lyapunov argument and moreover, we shall present that such a global attractor is indeed the set of optimal policy. In the following lemma, we  identify the Lyapunov function associated with \eqref{emp_freq_diff}

\begin{lemma}\label{le:vec_field}
	For every $s\in \mathcal{S}$ and any fixed $\omega(s)$ in $\Omega^{\eta'}$, let $d{\pi}_t(s)/dt=\omega(s)[\hat{\pi}_t(s)-\pi_t(s)], \hat{\pi}_t(s)=\mathcal{RM}(\pi_t(s),Q^{\pi_t}(s))$, then $\left\langle\nabla_{\pi_t} V^{\pi_t}(s), {d}{\pi_t}(s)/{dt} \right\rangle \geq 0.$
\end{lemma}



With this lemma, we now construct the Lyapunov function for \eqref{emp_freq_diff}, which further leads to the global convergence of the algorithm.  First, given $\pi^*$, an optimal policy, we define $L(\pi)=\sum_{s \in \mathcal{S}}\big[V^{\pi^*}(s)-V^{\pi}(s)\big].$
Apparently $L(\pi)$ is a positive semi-definite function, since the optimality gives $V^{\pi^*}(s)-V^\pi(s)\geq 0$ for all $s\in \mathcal{S}$ and $L(\pi)=0$ only if $\pi$ is an optimal policy. Then with \cref{le:vec_field}, for any $t>0$, we have 
	\begin{align*}
		\left\langle \nabla_{\pi_t}L(\pi_t), {d}{\pi_t}/{dt}\right\rangle=-\sum_{s\in\mathcal{S}}\left\langle\nabla_{\pi_t}V^{\pi_t}(s),{d}{\pi_t}(s)/{dt} \right\rangle\leq 0.
	\end{align*} 
This implies that $L(\pi)$ is a Lyapunov function for the differential inclusion \eqref{emp_freq_diff}, with a global attractor $\Pi=\{\pi: \pi \text{ is an optimal strategy}\}$, showing that $\pi_t$ given by \eqref{emp_freq_diff} converges almost surely to the attractor. Therefore, from the convergence result of the continuous dynamics, we claim the convergence of the coupled dynamics \eqref{onpolicy}, \eqref{emp_freq}.

\begin{proposition}
    	The sequence $\{Q_k,\pi_k\}_{k}$ given by the coupled recursive scheme \eqref{onpolicy} and \eqref{emp_freq} converges almost surely to $(Q^{\pi*}, \pi^*)$, where $\pi^*$ is an optimal policy and $Q^{\pi*}$ is associate optimal $Q$ values.
\end{proposition}



   
\section{Numerical Experiments}\label{sec:num}
In this section, we present experimental results when applying our Blackwell $Q$-learning to MDP problems.  Since our proposed method resembles expected SARSA \cite{seijen09SARSA}, we consider cliff walking task in that paper, where the agent has to find its way from the start to the goal in a grid world. The agent can take any of four-movement actions: up, down, left, and right, each of which moves the agent one square in the corresponding direction. Each step results in a reward of -1, except when the agent steps into the cliff area, which results in a reward of -100 and an immediate return to the start state. The episode ends upon reaching the goal state.

\begin{figure}[htp]
    \centering
    \includegraphics[width=0.42\textwidth]{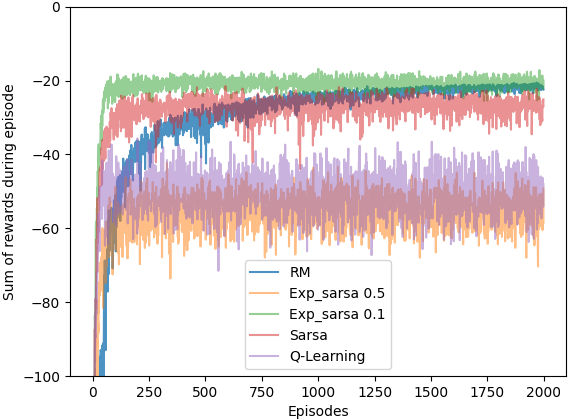}
    \caption{Comparison between different learning methods in cliff walking experiments: RM, Expected SARSA, SARSA and $Q$-learning.}
    \label{fig:compare}
\end{figure}

We evaluate the performance of $Q$-learning, SARSA, expected SARSA, and our Blackwell $Q$-learning. It is noted that our Blackwell $Q$-learning does not need any hyperparameter for encouraging exploration, as \eqref{eq:rm} always retains some probabilities for actions that yield positive regret. Hence, our method is less aggressive in terms of exploitation, compared with the others.  In our experiments, we adopt $\epsilon$-greedy policy for the first three, where $\epsilon=0.1$ for $Q$-learning and SARSA, and for expected SARSA, we run the algorithm with two different exploration rates $\epsilon=0.1, 0.5$.  We test these algorithms with 2000 episodes and we average the results over 200 independent runs. The numerical results are shown in Fig.~\ref{fig:compare}.
At first glance, both expected SARSA with $\epsilon=0.1$ and our Blackwell $Q$-learning give the best performance in the end, though the expected SARSA converges faster, due to the greedy policy. However, we note that the success of the expected SARSA relies on a carefully crafted exploration rate. If we set $\epsilon=0.5$, then the performance is even worse than that of SARSA. This observation highlights one merit of Blackwell $Q$-learning: it is hyperparameter free for exploration. 

Though in our experiments, Blackwell $Q$-learning seems not to outperform expected SARSA in terms of the convergence rate, because of the difference in action selection, we argue that such conservative action selection is actually more desired for online learning problem, where the environment is non-stationary. One prominent example is learning in games \cite{fudenberg_learning}, where the payoff is jointly determined by all players' actions. In this case, if one player only seeks the best response based on his own $Q$ function, he may not achieve any equilibrium in the end, as observed in \cite{nips2005_2834}. Due to the limited space, we fully develop our arguments in the supplementary.  
\section{Conclusion}\label{sec:conclusion}
We have introduced a novel approach for tackling MDP problems based on the Blackwell approachability theory. By constructing an auxiliary Blackwell game, we use its geometric interpretation to solve MDP problems by deriving no-regret learning from the Blackwell strategy, which provides an alternative to dynamic programming for MDP. Specifically, we have discussed one simple Blackwell strategy, regret matching, and how it can be incorporated into both offline planning methods (e.g., Blackwell value iteration) and online learning schemes (e.g. Blackwell $Q$-learning). Both are provably convergent. Related numerical results have been used  to corroborate our results. As for future work, we would like to extend our Blackwell approachability-based idea to online (adversarial) MDP \cite{mansour09onlinemdp} and multi-agent systems, where the environment is not stationary from any player's perspective, hence imposing difficulties on applying dynamic programming.   
\bibliographystyle{ieeetr}
\bibliography{ref}
\end{document}